\documentclass[preprint,12pt]{elsarticle}
\usepackage{amssymb}
\usepackage{amsmath}
\usepackage[numbers]{natbib}
\usepackage{graphicx} 
\usepackage{xurl}
\usepackage{algorithmic}
\usepackage{algorithm}
\usepackage{diagbox}
\usepackage{bm}
\usepackage{footmisc}
\usepackage{hyperref}
\usepackage{booktabs}

\makeatletter
\def\ps@pprintTitle{%
 \let\@oddhead\@empty
 \let\@evenhead\@empty
 \def\@oddfoot{\centerline{\thepage}}%
 \let\@evenfoot\@oddfoot}
\makeatother

\begin{document}

\begin{frontmatter}

\title{A Collaborative Process Parameter Recommender System for Fleets of Networked Manufacturing Machines -- with Application to 3D Printing} 

\author[1]{Weishi Wang\fnref{fn1}}
\ead{weishiw@umich.edu}
\author[2]{Sicong Guo\fnref{fn1}}
\ead{stevengu@umich.edu}
\author[2]{Chenhuan Jiang}
\ead{chenhj@umich.edu}
\author[4]{Mohamed Elidrisi}
\ead{melidris@cisco.com}
\author[4]{Myungjin Lee}
\ead{myungjle@cisco.com}
\author[3]{Harsha V. Madhyastha}
\ead{madhyast@usc.edu}
\author[1]{Raed Al Kontar}
\ead{alkontar@umich.edu}
\author[2]{Chinedum E. Okwudire\corref{cor1}}
\ead{okwudire@umich.edu}
\cortext[cor1]{Corresponding author(s)}
\fntext[fn1]{These authors contributed equally to this work as lead authors.} 

\affiliation[1]{organization={Department of Industrial and Operations Engineering, University of Michigan, Ann Arbor},
                country={U.S.}}

\affiliation[2]{organization={Department of Mechanical Engineering, University of Michigan, Ann Arbor},
                country={U.S.}}


\affiliation[3]{organization={Thomas Lord Department of Computer Science, University of Southern California, Los Angeles},
                country={U.S.}}
                
\affiliation[4]{organization={Cisco Systems, San Jose},
                country={U.S.}}

\begin{abstract}
Fleets of networked manufacturing machines of the same type, that are collocated or geographically distributed, are growing in popularity. An excellent example is the rise of 3D printing farms, which consist of multiple networked 3D printers operating in parallel, enabling faster production and efficient mass customization. However, optimizing process parameters across a fleet of manufacturing machines, even of the same type, remains a challenge due to machine-to-machine variability. Traditional trial-and-error approaches are inefficient, requiring extensive testing to determine optimal process parameters for an entire fleet. In this work, we introduce a machine learning-based collaborative recommender system that optimizes process parameters for each machine in a fleet by modeling the problem as a sequential matrix completion task. Our approach leverages spectral clustering and alternating least squares to iteratively refine parameter predictions, enabling real-time collaboration among the machines in a fleet while minimizing the number of experimental trials. We validate our method using a mini 3D printing farm consisting of ten 3D printers for which we optimize acceleration and speed settings to maximize print quality and productivity. Our approach achieves significantly faster convergence to optimal process parameters compared to non-collaborative matrix completion.
\end{abstract}


\end{frontmatter}
\section{Introduction} 
\par
\begin{figure}[htbp!]
    \centering
    \includegraphics[width=0.4\textwidth]{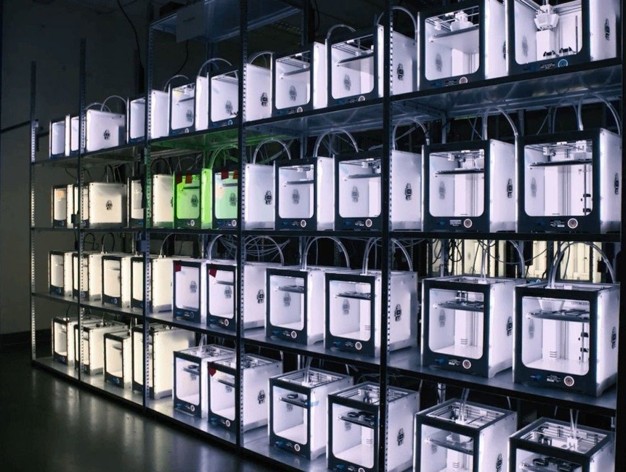}
	\caption{Example of a 3D printing farm. Such farms could have thousands of 3D printers working 24/7 to produce parts \cite{ultimakeryoutube}.}
	\label{Fig:ex_print_farm}
    \vspace{-2em}
\end{figure}

 Manufacturing firms increasingly deploy fleets of machines (e.g., machine tools, industrial robots, or 3D printers) of the same type (i.e., the same make and model) that are connected using a computer network \cite{michelsen2021changing}. The machines could be collocated or geographically dispersed. A prototypical example of this paradigm is 3D printing farms, which consist of multiple networked 3D printers operating in parallel, enabling faster production, improved redundancy, and efficient mass customization \cite{Skawinski2017, Weflen2021} -- see Fig.~\ref{Fig:ex_print_farm}. 
The printers often belong to the same manufacturing firm; therefore they share the same computer network. For instance, Prusa Research, based in the Czech Republic, operates a 3D printing farm of more than 600 3D printers for the company's in-house production \cite{All3DP2024}. JinQi Toys, based in China, runs a farm of 2500 printers for commercially producing sundry toys \cite{Schwaar2024}, while Slant3D, based in the United States, runs a farm of 800 printers for producing over 10,000 different parts per week for various customers. It has launched a plan to grow the size of its farm to over 3,000 printers in the near future \cite{All3DP2024}. 

\par However, even when a fleet consists of machines of the same type, performance differences arise due to variations in the machines' mechanical behaviors, leading to significant differences in the machines' production quality and efficiency \cite{Balta2018, Moylan2015, stoop2023cloud}.  
Furthermore, as machines age and undergo wear and tear, their performance gradually diverges, even under identical manufacturing conditions \cite{stoop2023cloud}. 
Machine-to-machine variability presents a major challenge in determining optimal process parameters (e.g., machine speed) across an entire fleet. Traditionally, operators rely on trial-and-error methods to iteratively test different parameter configurations to identify optimal process parameters \cite{Santana2017, Wang2018}. This approach is time-consuming, inefficient, and impractical for large-scale production, e.g., fleets with hundreds or thousands of machines. Without an efficient way to determine process parameters, operators are forced to apply the same parameters to all machines in a fleet, leading to suboptimal performance for individual machines.
\par To address the challenges of optimizing process parameters in additive manufacturing (AM) and other manufacturing processes, researchers have explored machine learning (ML) techniques \cite{Mahmood2021, wang2020101538}. For example, Dharmadhikar et al. \cite{Dharmadhikari2023} proposed a model-free reinforcement learning framework based on Q-learning to find optimal combinations of laser power and scan speed to maintain steady-state melt pool depth in laser-directed energy deposition AM. Chen et al. \cite{Chen2015} developed a neural network-based process parameter recommender system to predict part quality and printing time in binder jetting AM. Toprak et al. \cite{Toprak2025} proposed a ML model to predict the mechanical properties of parts produced using laser powder bed fusion AM. They then applied sequential quadratic programming to their ML model to determine optimal process parameters. In the arena of CNC machining, neural networks have been employed to correlate and optimize process parameters (like feedrate, cutting speed and depth) with respect to objectives (like energy consumption, surface roughness, and production rate) \cite{Wu2022, Zhang2024}.  However, the aforementioned ML-based approaches treat each machine as an independent system, requiring separate parameter optimization for every machine. They do not exploit collaborative learning across a fleet of manufacturing machines, which could significantly reduce the number of trials required to find optimal settings for each machine.
\par Researchers have investigated collaborative learning approaches in distributed manufacturing (DM),  which involves the use of a network of geographically dispersed machines that can share data to improve production efficiency \cite{DM2021, kontar2021internet}. For example, collaborative software agents have been incorporated into DM systems to address the problem of scheduling manufacturing resources \cite{Valilai2013, Shen2007}. 
A neural genetic algorithm has been used to seek solutions to resource allocation and production planning problems in DM \cite{Bootaki2024}.
Similarly, multi-agent reinforcement learning has been applied to resource allocation and production scheduling in smart factories, allowing machines to coordinate tasks and optimize workflows collectively \cite{Kim2020, Bahrpeyma2022}.  Reinforcement learning has also been employed to facilitate cloud-edge task offloading to improve production efficiency \cite{Guo2024}, and to enhance cooperative decision making \cite{Yuwono2024} in DM. 
Federated learning frameworks have been proposed to realize part geometry predictions and part qualification in distributed AM, 
where multiple machines contribute to a shared learning model while keeping their data local for privacy \cite{Mehta2024}. 
In cloud-based AM, researchers have explored the use of recommender systems to determine the optimal pipeline of manufacturing methods and solutions from available options \cite{Chen, Simeone, Liu}. 
Although the aforementioned studies highlight the value of collaborative intelligence in DM, no existing methods have applied such collaborative learning techniques to process parameter optimization for fleets of networked manufacturing machines.  
\par This paper seeks to bridge this gap in the literature and practice by introducing a machine learning-based collaborative process parameter recommender system for fleets of networked manufacturing machines, using 3D printing farms as a specific example. Unlike conventional approaches that optimize each machine's process parameters separately, our method enables the machines to collaborate by centrally sharing process data, allowing for more efficient and adaptive parameter selection across the fleet. The underlying hypothesis is that while the machines in a fleet exhibit variability, 
they also share enough structural and operational similarities that can be leveraged through collaborative learning to determine optimal process parameters more efficiently than traditional independent optimization.
To test this hypothesis, using the context of 3D printing farms, our paper makes the following original contributions:
\begin{enumerate}
    \item It formulates process parameter optimization across a fleet of networked manufacturing machines as a sequential matrix completion (MC) problem, enabling the machines in the fleet to collaborate through shared knowledge while maintaining machine-specific optimizations.
    \item It develops a scalable algorithm that integrates spectral clustering and alternating least squares (ALS) to iteratively refine parameter recommendations across a fleet of networked manufacturing machines.
    \item It demonstrates the effectiveness of the proposed recommender system through experiments on a mini 3D printing farm with ten printers, showing that the proposed collaborative MC approach achieves faster convergence to optimal printing conditions compared to a non-collaborative MC technique.
\end{enumerate}

The remainder of this paper is organized as follows: 
Sec.~\ref{Sec:methods} introduces the proposed method, detailing the sequential matrix completion framework and the collaborative learning algorithm. Sec.~\ref{Sec:exp} presents the experimental setup, including details of the 3D printers, process parameters, and evaluation criteria. Sec.~\ref{Sec:results} discusses the results, comparing the performance of our method against a baseline non-collaborative technique. Finally, Sec.~\ref{Sec:conclusion} provides conclusions and outlines directions for future research.

\section{Methodology} \label{Sec:methods}
\subsection{Problem Setup}
We consider a fleet of $K$ networked manufacturing machines, each with $d$ process parameters that assume discrete values. The discrete values for each process parameter can be represented as an ordinal set $\mathcal{X}_{i}=\{1,2,\cdots,m_i\}$,  for $i\in\{1,2,\cdots,d\}$, and the entire process parameter space for each machine is given by the Cartesian product $\mathcal{X}=\mathcal{X}_{1}\times \mathcal{X}_{2}\times\cdots\times \mathcal{X}_{d}$. Our objective is to determine the optimal operating condition for each machine that maximizes an output $u(\bm{x}): \mathcal{X} \rightarrow \mathbb{R}$ where $\bm{x}\in\mathcal{X}$ is the operating process parameter. 

The output can be defined based on a single performance metric, such as manufacturing quality, or a weighted combination of metrics, such as quality and efficiency. The primary challenge in optimizing process parameters arises from the black-box nature of the problem. Understanding the effect of a process parameter \( \bm{x} \in \mathcal{X} \) on the output \( u(\bm{x}) \) requires running experiments. However, conducting exhaustive experiments to determine the optimal operating conditions for each machine in the fleet is both time-consuming and costly, particularly when the size of the fleet and the parameter space are large (i.e., when $K$, \( d \) or \( m_i \) are large) and experimental evaluations are expensive. This raises the question of how to efficiently explore the parameter space while minimizing excessive experimentation for each individual machine.

More critically, machines may exhibit heterogeneous mechanical characteristics, meaning that they do not necessarily respond identically to the same operating conditions. As such, instead of finding an optimal set of the process parameters $\bm{x}^{\star}$ across all machines, one may need to find an individualized optimal condition $\bm{x}^{\star}_k$ for each machine $k \in [1, \cdots, K]$. 

\par Collaboration provides a crucial solution to this challenge. While machines may differ in behavior, they often share commonalities in performance trends. By strategically borrowing strength from others, we can significantly reduce the need for extensive individual experimentation. Instead of treating each machine as an independent entity, we leverage shared knowledge across similar machines, accelerating the learning process while still allowing for machine-specific customization. This collaborative approach minimizes the experimental burden and improves efficiency in determining optimal operating conditions across the fleet.

\par Therefore, our goal is to efficiently recommend the best operating conditions for each machine via collaborative sequential experimentation, leveraging both similarities and differences among machines to accelerate optimization while reducing experimentation costs.

\subsection{Proposed Approach}
\label{sec:seq_comp_fml}

\subsubsection{Matrix Completion for Collaborative Learning}
\label{sec:mc}

Assuming all machines operate under a shared process parameter space, how can we exploit shared knowledge?

\par
\begin{figure*}
	\centering
	\includegraphics[width=.9\textwidth]{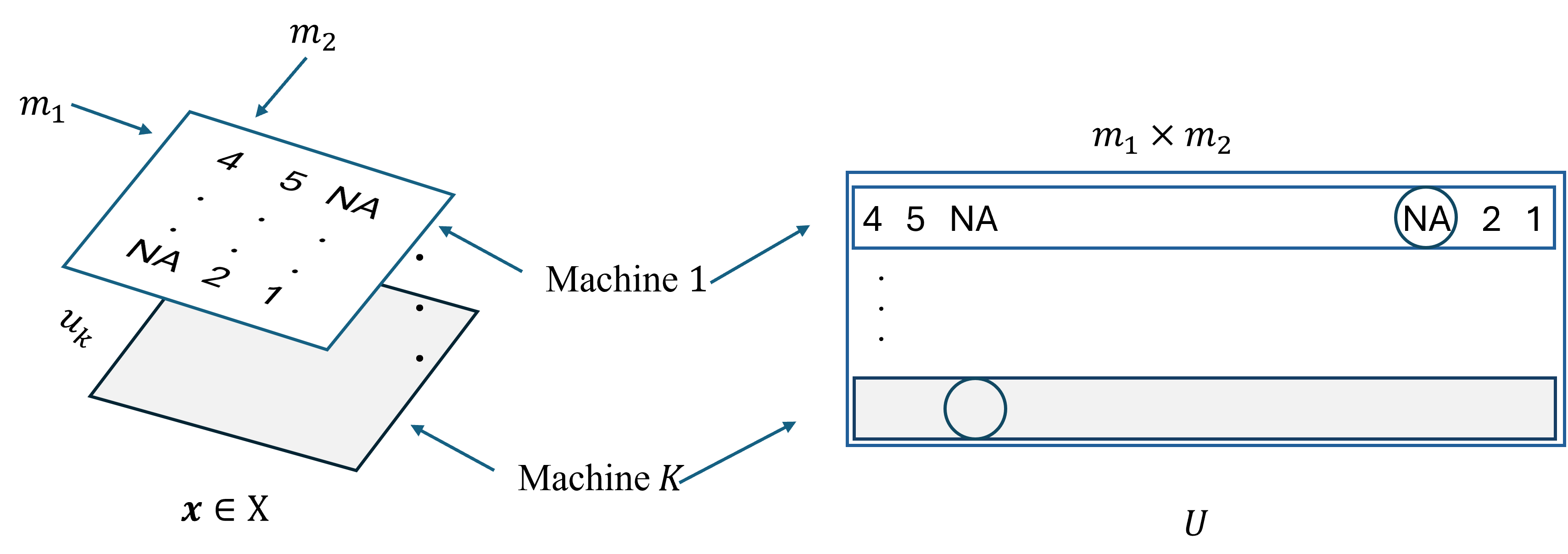}
	\caption{Squeezing operating tensors $u_{k}$ into vectors $\text{Vec}(u_{k})$ and stacking them to form the collaborative matrix $U$. $\mathbf{x}\in\mathcal{X}$ is the process parameter we need to optimize over. The blue circle is the optimal location $\mathbf{x}^{\star}_{k}$ for each machine. The entry with the value 'NA' denotes the missing entry.  We assume $d=2$ here for clear visualization.}
	\label{FIG:squeeze_stack}
\end{figure*}

 To formalize this problem, consider the case where we have two process parameters, i.e., $d=2$, then the output function for each machine $k$ can be represented as a matrix $u_{k}(\mathcal{X})\in \mathbb{R}^{m_{1}\times m_{2}}$ (see Fig. \ref{FIG:squeeze_stack}). When $d>2$, the output naturally extends to a high-dimensional tensor $ u_{k}(\mathcal{X}) \in \mathbb{R}^{m_1 \times m_2 \times \cdots \times m_{d}} $. This representation captures the relationship between the values of different process parameters and the corresponding performance metrics.
\par To facilitate collaboration, we propose a MC framework that structures the exploration space in terms of a sparse matrix across all $K$ machines that need to be fully learned. Specifically, we vectorize the tensor $u_{k}$ of each machine into a vector representation:
\begin{equation}\label{eq:vectorize1}
    \text{Vec}(u_{k}) \in\mathbb{R}^l, \;\text{where}\; l=m_{1}\cdot m_{2}\cdots \cdot m_{d}. 
\end{equation}
Then we aggregate the vectors to form the collaborative matrix $U\in\mathbb{R}^{K\times l}$ where missing entries correspond to unobserved or unexplored process parameter values for each machine. Our objective is to recover the missing entries in $U$ via MC, leveraging similarities across machines to enhance learning efficiency. The overall process is illustrated in Fig.~\ref{FIG:squeeze_stack}.

\par Indeed, MC has garnered significant attention over the past decade, especially within recommender systems, which have been widely deployed in various industries, including e-commerce and entertainment (e.g., Netflix) \cite{Koren2009}. 

More specifically, we assume the data generation model \cite{nguyen2019low} over $U$ as 
\begin{align}
    U=AB^{\top} + E, 
    \label{eq: generation_model_large}
\end{align}
where $A\in\mathbb{R}^{K\times r}, B\in\mathbb{R}^{l\times r}$ are two low-dimensional factor matrices of rank $r < \min(K,l)$ capturing the latent structure matrices of $U$, and $E$ denotes additive noise. Notice that in (\ref{eq: generation_model_large}), the matrix \( A \) encodes machine-specific coefficients, with each row \( A_{k, \cdot} \in \mathbb{R}^{1 \times r} \) representing a unique machine and its latent response to process parameters. Meanwhile, \( B \) captures the underlying structure of the process parameter space, serving as a shared basis across all machines. The interaction between \( A \) and \( B \) allows each machine to weight the global process trends differently, enabling the model to learn individualized responses while leveraging collective knowledge. 

Now recall that \( U \) is a matrix with only a few known entries. To define our loss function, we denote \( \Omega \) as the support set containing the indices of the observed entries in \( U \). Assuming that the rank \( r \) is known (to be discussed in Sec.~\ref{sec:practical_consid}), our objective can be written as:

\begin{align}
    \min_{A,B}\frac{1}{2}&\|\mathcal{P}_{\Omega}(U-AB^{\top})\|_{F}^2 \notag \\ & +\lambda\left(\sum_{j=1}^{r}\|A_{\cdot, j}\|_{2}^{2}+\|B_{\cdot, j}\|^{2}_{2}\right), 
    \label{eq:loss}
\end{align}
where $\mathcal{P}_{\Omega}$ is the projection over the observations $\Omega$ (i.e., the non-missing entries in $U$), $\lambda$ is a regularization parameter and $A_{\cdot, j}$ refers to the $j$-th column of $A$. The first term minimizes the reconstruction error of \( U \) based on the observed entries in the support set \( \Omega \). The second term acts as a regularization penalty to prevent overfitting.

Fortunately, this class of problems can be efficiently solved using alternating least squares (ALS), where we fix \( A \) and optimize \( B \), then alternate by fixing \( B \) and optimizing \( A \). Each step simply solves a ridge regression problem, which has a closed-form solution. The details of this approach can be found in \cite{nguyen2019low, zhou2008large}. Such an ALS approach can also be easily parallelized for large-scale implementations.  

After obtaining the estimated low-rank matrices \( \widehat{A} \) and \( \widehat{B} \) (matrices $\widehat{A},\widehat{B}$ are estimation of matrices $A$ and $B$), we reconstruct \( U \) by predicting the missing entries as \( \widehat{U} = \widehat{A} \widehat{B}^{\top} \) where the matrix $\widehat{U}$ is the estimation of the matrix $U$.

To ensure identifiability and obtain a feasible initialization for the matrix factorization, it is necessary that each machine (i.e., each row in the collaborative matrix $U$) has at least one observed entry. However, even with this minimal condition satisfied, the overall matrix $U$ remains extremely sparse in the early stages of experimentation. Such sparsity poses a significant challenge, as it may prevent the model from capturing meaningful global patterns or accurately recovering individual machines' behaviors. To address this issue, we introduce a sequential matrix completion framework in the next section, which adaptively selects new process parameters to query. This enables more efficient exploration of the process parameter space $\mathcal{X}$, accelerates the convergence of matrix recovery, and ultimately reduces the cost of experimentation.

\subsubsection{Sequential Matrix Completion}
\label{sec: seq_matrix_completion}

\par While the initial matrix contains sparse entries, additional experimentation can be strategically conducted to uncover missing entries in \( U \) and refine the MC process. The fundamental challenge is determining which entries to uncover to maximize the overall effectiveness of MC. To address this, we propose a sequential MC framework that iteratively selects new process parameter settings to explore.

Hereafter, we refer to \( U \) as our utility matrix, as it quantifies the benefit obtained from each process parameter value \( \bm{x} \). We assume a fixed budget of \( M \), representing the number of new observations that can be sequentially gathered. The next step is to determine where to sample for the next experiment (or batch). A natural approach is to select unobserved entries that have the potential to maximize \( U \). To this end, we first use MC to fully predict \( \widehat{U} \), and then consider two scenarios:

\begin{itemize}

\item \textbf{Full Fleet Participation}. If all $K$ machines in the fleet can participate in collaborative experimentation, the next experiment for machine \( k \) is selected by first identifying the column \( j^\star_k \) in the \( k \)-th row of the utility matrix \( \widehat{U} \) that maximizes the predicted utility:
\begin{align}
    j^{\star}_{k} = \arg\max_{j} \widehat{U}_{k,j}.
    \label{eq:select1}
\end{align}
The process parameter values \( \bm{x}_{j^\star_k} \) corresponding to this column are then selected for the next experiment. This results in \( K \) independent experiments, allowing each machine to explore its most promising process parameter values while benefiting from shared knowledge across the fleet.
\item \textbf{Limited Fleet Participation}. If the participation is restricted such that only a subset \( c \) of the \( K \) machines in the fleet can conduct experiments at any given time, we must prioritize the most informative ones. In this case, we first select the top \( c \) machines whose rows \( \widehat{U} \) contain the maximum values:
\begin{align*}
     \{k_{1}^{\star}, k_{2}^{\star}, \dots, k_{c}^{\star} \} = \text{Top}_c \left(\max_{j} \widehat{U}_{k,j} \right),   
\end{align*}
where \( \text{Top}_c(\cdot) \) selects the \( c \) largest values in descending order. Next, for each of these selected machines, we determine the process condition \( \bm{x}_{j^\star_k} \) corresponding to the column that maximizes its predicted utility:
\begin{align}
    j^{\star}_{k} = \arg\max_{j} \widehat{U}_{k,j}, \, \forall k \in \{k_{1}^{\star}, k_{2}^{\star}, \dots, k_{c}^{\star} \}.
    \label{eq:select2}
\end{align}
This ensures that the most promising process parameter values are explored while efficiently utilizing the limited available resources. 
\end{itemize}

\subsection{Overall Algorithm}
\label{sec:overall_algorithm}

By combining the elements of MC and sequential selection, we establish an overall framework for optimizing process parameters across multiple machines. This framework iteratively refines manufacturing conditions by leveraging information sharing and guided experimentation. The complete procedure is summarized in Algorithm~\ref{alg: CDAM}.

\begin{algorithm}[ht!]
\caption{Overall Framework}
\label{alg: CDAM}
\begin{algorithmic}[1]
\STATE \textbf{Input}: Initial observed set $\Omega^{(0)}$, $M$, estimated rank $r$ of $U$.
\STATE \textbf{Initialize} Define $u_k$ using (\ref{eq:vectorize1}) and form $U^{(0)}$.
\FOR{$t = 1$ \TO $M$}
    \STATE \textbf{Matrix Completion}: Learn  $\widehat{U}^{(t)}$ from (\ref{eq:loss}).
    \STATE \textbf{Select experiments}: Select  $\bm{x}^{(t)}_{j^\star_k}$ using (\ref{eq:select1}) or (\ref{eq:select2}).
    \STATE \textbf{Experiment}: Conduct experiments at $\bm{x}^{(t)}_{j^\star_k}$ 
    \STATE \textbf{Update Data}: Augment $U^{(t)}$ with new data to form $U^{(t+1)}$, and obtain $\Omega^{(t+1)}$.
\ENDFOR
\STATE \textbf{Return}: Optimal process parameters $\bm{x}^{(M)}_{j^\star_k}$ for each machine $k$.
\STATE \textbf{End Algorithm}
\end{algorithmic}
\end{algorithm}

\par The proposed framework follows two key iterative steps:

\begin{itemize}
    \item \textbf{Matrix Completion}. We model the utility matrix \( U \) as a low-rank structure and estimate missing values. This approach leverages shared trends across machines to iteratively refine the predictions of unexplored manufacturing conditions.
    \item \textbf{Candidate Selection for Sequential Experimentation}. We select and test the most informative process conditions. This selection follows a utility-driven sampling strategy that selects high-potential parameters, while considering budget constraints. The observed data is updated iteratively to further improve the MC process.
\end{itemize}

\par By iterating these steps, our framework efficiently optimizes process parameters while minimizing experimental costs.

\subsection{Practical Considerations}
\label{sec:practical_consid}
A requisite step in our approach is estimating the rank of the collaborative utility matrix \( U \). Since each row of \( U \) represents an individual machine and the number of process conditions far exceeds the number of machines, the rank estimation process can be interpreted as identifying clusters of machines with similar operational behaviors. We achieve this through a combination of matrix imputation and clustering.

We leverage column-wise information through imputation to address missing entries in $U$. Various methods can be employed, including mean imputation \cite{donders2006gentle}, multivariate imputation by chained equations (MICE) \cite{van2011mice}, and heterogeneous matrix completion \cite{shi2023heterogeneous}. In our implementation, we adopt mean imputation, replacing missing values in $U$ with their respective column means. This preprocessing step enhances data completeness, enabling robust clustering for rank estimation. Specifically, we apply spectral clustering \cite{von2007tutorial}, constructing an affinity matrix using a Gaussian kernel to capture machine similarities. By clustering machines with analogous operational characteristics, we obtain a principled rank estimate that effectively captures the underlying structure of $U$.

\section{Application to 3D Printing Farm} \label{Sec:exp}

In fused filament fabrication (FFF) 3D printing, mechanical vibrations are a critical factor influencing both print quality and process efficiency (productivity). These vibrations, which often occur during rapid changes in direction or speed, can leave visible artifacts on printed parts, commonly referred to as ringing or ghosting \cite{duan2018limited}. Ringing typically appears as oscillatory patterns near sharp corners. While increasing speed and acceleration improves productivity by reducing print times, it typically intensifies ringing, leading to poorer surface finish. Conversely, lowering these settings reduces ringing but at the cost of longer print durations and reduced throughput.

This tradeoff between quality and productivity is further complicated by machine-to-machine variability, even among printers of the same make and model. Factors such as differences in initial assembly, tolerances in mechanical components, and long-term wear, such as loosening belts and bolts, can cause printers to exhibit different vibration profiles. As a result, the optimal combination of speed and acceleration that balances quality and productivity can vary from one printer to another. This variability makes it an ideal scenario to evaluate our proposed collaborative matrix completion (MC) framework, which aims to identify machine-specific optimal process parameters while leveraging shared knowledge across a networked fleet.

\subsection{Experimental setup and procedure}\label{sec:exp_setup}

\begin{figure*}
    \centering
    \includegraphics[width=\textwidth]{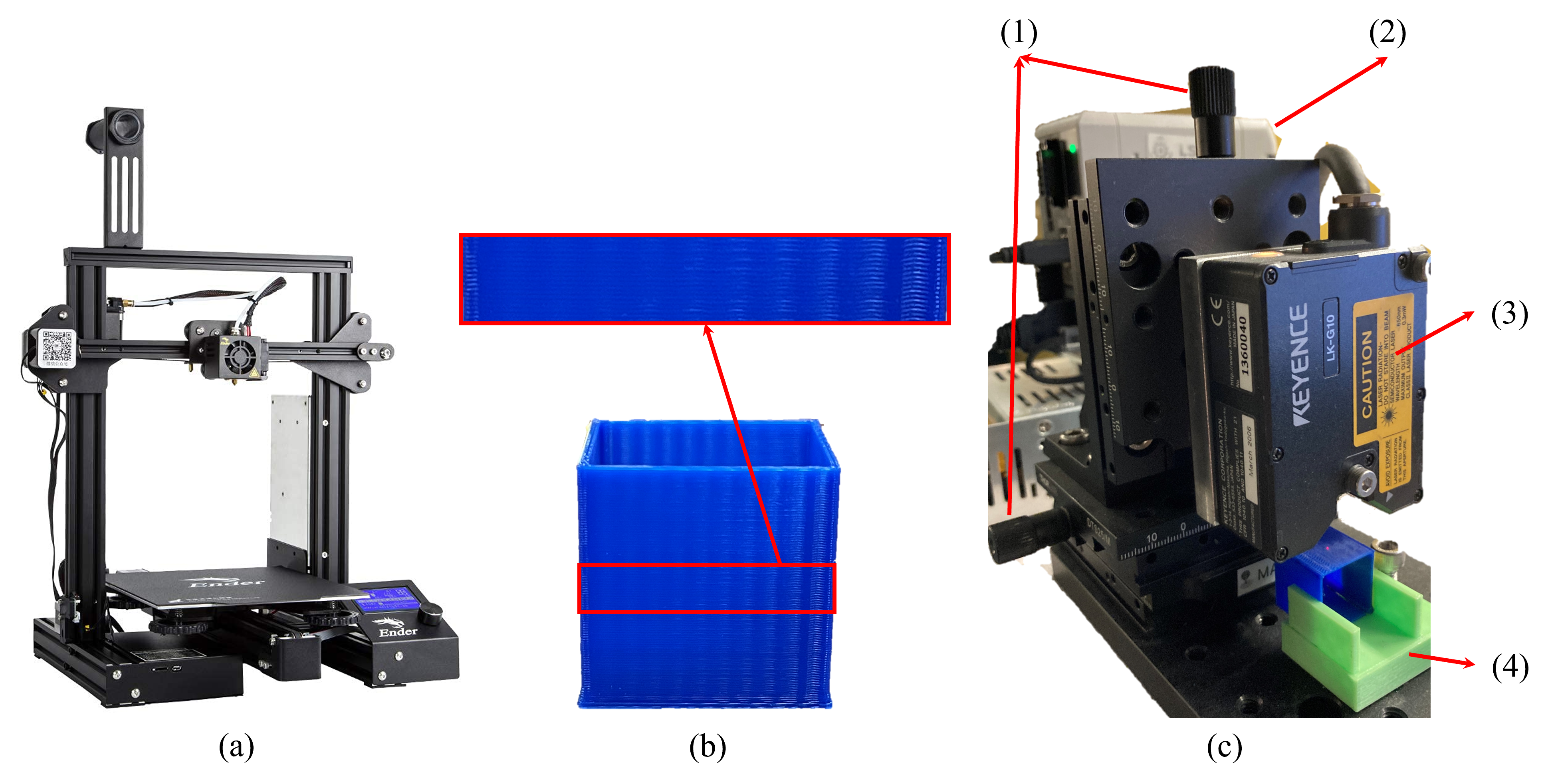}
	\caption{The experimental setup for Sec.~\ref{sec:exp_setup}. (a) Creality Ender-3 Pro desktop FFF 3D printer. (b) Topless cube with the scanned section highlighted by the red rectangular area and enlarged above to show the ringing defects. (c) Laser scanner setup consisting of: (1) Thorlabs DTS25/M motorized translation stage; (2) the Keyence LK-GD500 controller; (3) the Keyence LK-G3001 laser displacement sensor; and (4) the part holder.}
	\label{Fig:exp_setup}
\end{figure*}

The experimental testbed is a mini 3D printing farm consisting of $K = 10$ Creality Ender-3 Pro desktop FFF 3D printers (see example in Fig.~\ref{Fig:exp_setup}(a)). These printers are supplied by the manufacturer as do-it-yourself (DIY) kits, requiring end users to complete the assembly. In this study, each of the ten printers was assembled by a different user, introducing natural variability in alignment, tensioning, and other mechanical characteristics. The printers have also been subjected to different levels of usage, which may have resulted in varying degrees of wear and tear. These mechanical variations are known to influence vibration behavior and are a source of printer-to-printer differences in print quality.

The print job involved printing of a topless cube measuring 20 × 20 × 20 mm$^3$, chosen for its simplicity and clear manifestation of vibration-induced defects (see Fig.~\ref{Fig:exp_setup}(b)). All prints were conducted in the same room for several days, using a HATCHBOX 1.75 mm True Blue PLA filament. Although minor variations in environmental conditions such as temperature and humidity may have occurred during the testing period, these factors are not considered to be the major contributors to the ringing defects that are the focus of this study. Ringing artifacts in FFF printing are primarily caused by inertial and mechanical factors rather than atmospheric ones.

Each of the Ender-3 Pro printers in the fleet was used to fabricate the aforementioned cube with varying combinations of two process parameters (i.e., $d =2$). They are: (maximum) printing speed ($v$) and (maximum) printing acceleration ($a$). Printing speeds were varied from 50 mm/s to 150 mm/s in increments of 25 mm/s (i.e., $m_{1} = 5$), and printing acceleration values ranged from 4000 mm/s$^{\text{2}}$ to 7000 mm/s$^{\text{2}}$ in increments of 500 mm/s$^{\text{2}}$ (i.e., $m_{2} = 7$), forming a grid of $l = m_{1}\times m_{2} = 35$ process parameter values. 

Therefore, with each combination of process parameters, a cube $n$ ($n=1,2,...,35$) was fabricated on each printer $k$ ($k=1,...,10$). As a measure of print efficiency (productivity), the printing time $\tau_{k,n}$ for each cube $n$, printed using printer $k$ was recorded from the start to the end of each print job. 
The print quality of each cube was assessed by measuring the surface roughness primarily caused by ringing. To do this, we utilized a Keyence LK-G3001 laser displacement sensor in conjunction with a LK-GD500 controller and a Thorlabs DTS25/M motorized translation stage with a part holder to clamp each cube while scanning its surfaces (see Fig.~\ref{Fig:exp_setup}(c)). This setup enabled high-resolution surface scanning of printed parts. The $x$ and $y$ surfaces of each cube were scanned across a $20$ mm wide and $4$ mm tall region in the middle of each surface (as indicated on Fig.~\ref{Fig:exp_setup}(b)), using a pitch of 1 mm. A slow and constant traverse speed of 3.5 mm/s was maintained during each scan. The sampling frequency of the scanner was 250 Hz, leading to $p = 7140$ grid points per surface (i.e., 1428 points per scan line times 5 scan lines per surface). 

Let $S_{x,k,n}\in\mathbb{R}^{p \times 1}$ and $S_{y,k,n}\in\mathbb{R}^{p \times 1}$ represent the raw data acquired from scanning the $x$ and $y$ surfaces of each cube $n$ printed using printer $k$. 
The quality (roughness) of $x$ and $y$ surfaces of each cube $n$ printed with printer $k$  was quantified as $q_{x,k,n}$ and $q_{y,k,n}$ given by: 

\begin{equation}\label{eq:quality_k}
\begin{aligned}
    q_{x,k,n}&=\sqrt{\frac{1}{p}\left(S_{x,k,n}^{\top}S_{x,k,n}-\frac{1}{p}\left(\boldsymbol{1}^{\top}S_{x,k,n}\right)^{2}\right)},\\
    q_{y,k,n}&=\sqrt{\frac{1}{p}\left(S_{y,k,n}^{\top}S_{y,k,n}-\frac{1}{p}\left(\boldsymbol{1}^{\top}S_{y,k,n}\right)^{2}\right)},
\end{aligned}
\end{equation}
where $\boldsymbol{1}\in \mathbb{R}^{p \times 1}$ is a column vector of ones. Eq. \eqref{eq:quality_k} represents the root mean square value of the data acquired from the scanned surfaces offset by the mean of the data. Combining $q_{x,k,n}$ and $q_{y,k,n}$, the quality  $q_{k,n}$ of cube $n$ printed by printer $k$ is expressed as:
\begin{equation}
    q_{k,n}=\sqrt{\frac{q_{x,k,n}^{2}+q_{y,k,n}^{2}}{2}} \quad\text{for} \quad n \in \{1, \cdots, l\}.
\end{equation}
The vectors $q_{k}\in\mathbb{R}^{l \times 1}$ and $\tau_{k}\in\mathbb{R}^{l\times 1}$, representing the print quality and print efficiency for each printer $k$ were then defined as
\begin{equation} \label{eq:p1} q_{k}=\begin{bmatrix}q_{k,1}\\q_{k,2}\\\vdots\\q_{k,l}\end{bmatrix},\quad \tau_{k}=\begin{bmatrix}\tau_{k,1}\\\tau_{k,2}\\\vdots\\\tau_{k,l}\end{bmatrix}.
\end{equation}

\subsection{Printing utility function} 
To enable multi-objective optimization in evaluating print outcomes, a unified \textit{printing utility} score was computed for each printer $k$ 
by combining $q_{k}$ and $\tau_{k}$ into a single vector, $u_{k}$.

To achieve this, both $q_k$ and $\tau_k$ were first normalized to $\bar{q}_k\in\mathbb{R}^{l \times 1}$ and $\bar{\tau}_k\in\mathbb{R}^{l \times 1}$, respectively, by dividing each vector by its maximum element, thereby ensuring that all entries lie within the range $[0, 1]$. We then defined a weighting function that adjusts the relative importance of print quality and efficiency based on the value of $\bar{q}_k$ as:
\begin{equation}
\label{eq:weight_definition}
    w_{q,k} = 0.78 \left(e^{0.82\bar{q}_{k}} - \boldsymbol{1} \right), \qquad w_{\tau,k} = \boldsymbol{1} - w_{q,k} \, ,
\end{equation}

\noindent where $w_{q,k}\in\mathbb{R}^{l\times 1}$ is the weight assigned to the normalized print quality $\bar{q}_{k}$ and $w_{\tau,k}\in\mathbb{R}^{l\times 1}$ is the complementary weight assigned to 
the normalized printing time $\bar{\tau}_k$.

This weighting scheme places increasing emphasis on print quality as surface roughness increases (i.e., $\bar{q}_{k}$ increases), thus penalizing poor-quality prints more heavily. Conversely, when surface quality is already high (i.e., $\bar{q}_{k}$ is low), the formulation gives greater weight to printing efficiency. 

\textbf{Remark}: \textit{The weighting formulation of \eqref{eq:weight_definition} is just an example. One can decide on other weighting formulations that align with desired outcomes.}

Therefore, for each printer $k$, the overall printing utility, $u_{k}$, was then computed 
with a weighted sum of the normalized quality $\bar{q}_{k}$ and print time $\bar{\tau}_{k}$ using Hadamard products:
\begin{equation}
\label{eq:utility_definition}
    u_{k} = -(w_{q,k} \odot \bar{q}_{k} + w_{\tau,k} \odot \bar{\tau}_{k} \,) .
\end{equation}

\noindent Note that the negative sign in \eqref{eq:utility_definition} indicates that higher values of $u_{k}$ (implying lower values of $\bar{q}_{k}$ and $\bar{\tau}_{k}$) lead to more favorable trade-offs between print quality and speed. This utility score is used as the target value to be maximized in the collaborative optimization framework described in Sec.~\ref{sec:overall_algorithm}.

\begin{figure*}
    \centering
    \includegraphics[width=\textwidth]{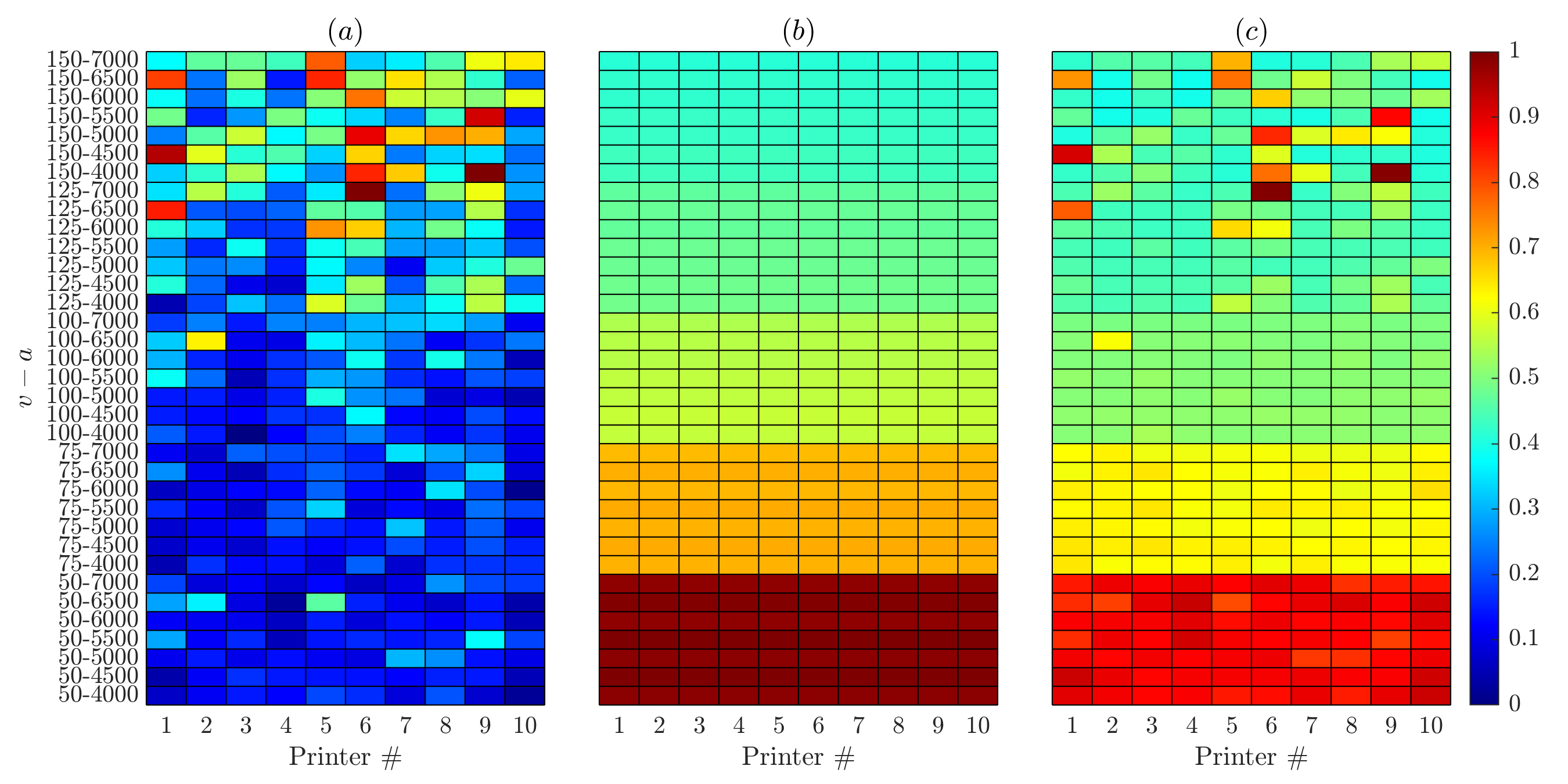}
	\caption{The investigated (a) normalized print quality $\bar{Q}^{\top}$; (b) normalized print time $\bar{T}^{\top}$; (c) absolute value of print utility $-U^{\top}$ of all configurations of printing speed $v$ (mm/s) - acceleration $a$ (mm/$\text{s}^{2}$) among all ten printers. Note that the negative sign applied to $U^{\top}$ allows it to be compared with $\bar{Q}^{\top}$ and $\bar{T}^{\top}$ using the same [0, 1] scale.}
	\label{Fig:full_Q}
\end{figure*}

\subsection{Implementation of recommender algorithm}
\label{sec:implement}

\par To evaluate the effectiveness of our proposed collaborative learning approach, we apply the sequential matrix completion algorithm described in Algorithm~\ref{alg: CDAM} under two practical scenarios: Case 1: Full Fleet Participation, where all $K$ printers in a fleet can participate in collaborative experimentation; and Case 2: Limited Fleet Participation, where only a subset $c$ of the $K$ printers in a fleet can participate in collaborative experimentation. 
\par We simulate an initial data collection scenario by introducing artificial sparsity into the utility matrix $U \in \mathbb{R}^{10 \times 35}$. Specifically, we randomly mask 192 (55\%) of the entries in $U$ (i.e., we retain 158 (45\%) as initially observed), reflecting a realistic setting where only a subset of process conditions have been explored across the fleet. The remaining entries are treated as missing and must be inferred through collaborative matrix completion. 

At the beginning of every iterative trial $t$ to exploring new process parameter settings, the utility matrix is estimated as $\widehat{U}^{(t)}$ via (\ref{eq:loss}), with the regularization parameter $\lambda=0.05$.

\par For Case 1, the total experimental budget is set to $M = 19$, representing the number of additional experiments (i.e., new parameter configurations) that can be tested across the fleet. We adopt a low-rank factorization model for MC, and the rank of the utility matrix $U$ is estimated as $r = 3$ based on the imputation and spectral clustering procedure outlined in Section~\ref{sec:practical_consid}. To benchmark performance, we also implement a non-collaborative variant of the algorithm in which each machine independently estimates its utility matrix $\hat{u}_{k}(\mathcal{X})$ using only its own observed entries by matrix completion. Each machine then selects the next process parameter configuration by solving $\text{argmax}_{\bm{x} \in \mathcal{X}} \hat{u}_{k}(\bm{x})$, without leveraging shared information from other machines.

\par For Case 2, we still set the total budget $M=19$, but we assume that only $c=5$ of the $K=10$ machines are available for experimentation in each round due to constraints such as time, material, or labor. In this case, a subset of the fleet is selected in each iteration according to the utility-based prioritization strategy described in Sec.~\ref{sec: seq_matrix_completion}, ensuring that the most rewarding process parameter settings are explored within the budget. As in Case 1, we compare against a non-collaborative baseline where $c=5$ machines are chosen randomly in each round and perform MC using only their own data, without sharing information across the fleet.


\section{Results \& Discussion} \label{Sec:results}

We evaluate the effectiveness of our proposed collaborative sequential MC method against a non-collaborative baseline for Cases 1 and 2 described in the preceding subsection. 

Acquired from the experiments described in Sec.~\ref{sec:exp_setup}, matrices of normalized print quality $\bar{Q}$, normalized print time $\bar{T}$, and print utility $U$ from all $K$ printers are constructed as \begin{equation}
    \bar{Q}=\begin{bmatrix}
        \bar{q}^{\top}_{1}\\\bar{q}^{\top}_{2}\\\vdots\\\bar{q}^{\top}_{K}
    \end{bmatrix},
    \quad \bar{T}=\begin{bmatrix}
        \bar{\tau}^{\top}_{1}\\\bar{\tau}^{\top}_{2}\\\vdots\\\bar{\tau}^{\top}_{K}
    \end{bmatrix},
    \quad U=\begin{bmatrix}
        u^{\top}_{1}\\u^{\top}_{2}\\\vdots\\u^{\top}_{K}
    \end{bmatrix} \, , 
\end{equation}
where $\bar{q}_{k}$, $\bar{\tau}_{k}$ and $u_{k}$ have been respectively defined in (\ref{eq:p1}) - (\ref{eq:utility_definition}). These matrices are illustrated in Fig.~\ref{Fig:full_Q}, where we plot the transpose of each matrix (and the negative of $U$) for ease of visualization. Notice the large variation in print quality $\bar{Q}$ among the ten printers, caused by significant differences in their vibration behaviors.

Two performance metrics are used to assess our proposed approach and benchmark:

(i) \textbf{Average number of trials to find optimal process parameters.} 
We define 
$t_k^\star = \min \left\{ t \in [1, M] \, | \, \bm{x}^{(t)}_{j^\star_k} = \bm{x}^{\star}_k \right\}$ 
as the number of trials, within the budget $M$, required to find the true optimal setting 
$\bm{x}^{\star}_k$ for each printer $k$. 
Note that $\bm{x}^{\star}_k$ is known, since we have access to the complete utility matrix $U$ but 
intentionally masked 55\% of its entries for evaluation purposes. 
We then report the average number of trials across all printers as 
$\frac{1}{K} \sum_{k=1}^{K} t_k^\star$.

(ii) \textbf{Cumulative regret for each printer.} 
At each trial $t$ within our algorithm, we define the regret for printer $k$ as  
\begin{equation*}
    \text{regret}^{(t)}_{k}= \left| u_k(\bm{x}^{(t)}_{j^\star_k}) - u_k(\bm{x}^{\star}_k) \right| = \left| \widehat{U}^{(t)}_{k,j_{k}^{\star}} - U_{k,j_{k}^{\text{true}}} \right| \, ,
\end{equation*}
where $j_{k}^{\text{true}} = \arg\max_{j} U_{k,j}$, based on the fully observed utility matrix $U$. 
Essentially, $\text{regret}^{(t)}_{k}$ quantifies the difference in utility between the true optimal process parameters for printer $k$ and the parameters selected by the algorithm at trial $t$. We then report the cumulative regret for each printer as $\sum_{t=1}^{M} \text{regret}^{(t)}_{k}$.

The first metric captures how quickly the algorithm can identify the true optimal process parameters for each printer. In contrast, cumulative regret, commonly used in online decision-making \cite{neu2010online}, offers a more comprehensive view of the algorithm's behavior across all $M$ trials. While the average number of trials is most useful when the only goal is eventually reaching the optimal setting, cumulative regret becomes especially relevant when suboptimal choices along the way have real consequences. For example, if printing with poor parameters produces defective parts that must be discarded, then simply finding the best setting in the end is not enough; \textit{every trial matters}. In such cases, cumulative regret highlights the quality of all decisions made during the process, not just the final outcome.

\begin{figure*}
	\centering
	\includegraphics[width=\textwidth]{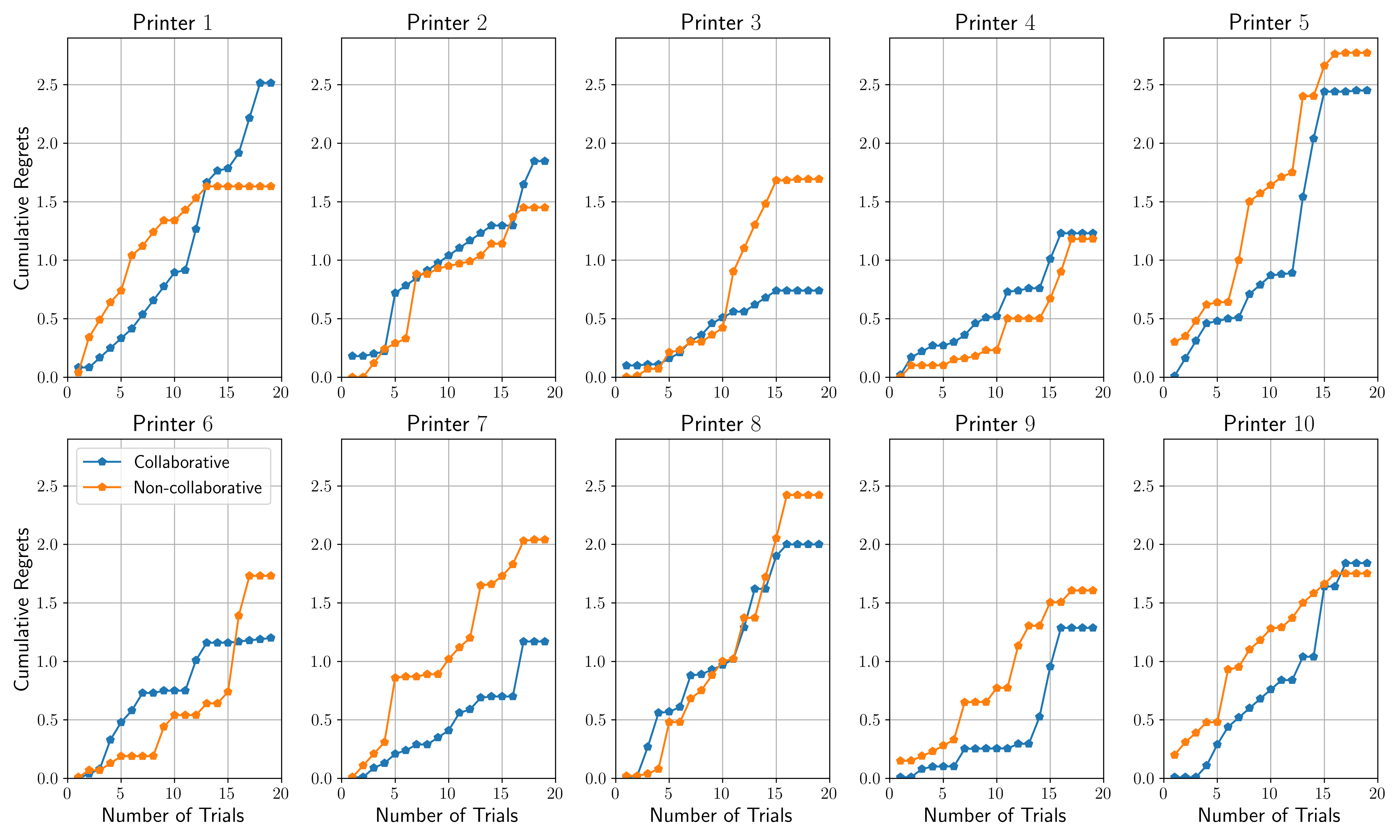}
    \caption{Cumulative regret of our proposed collaborative MC benchmarked against non-collaborative MC over $K=10$ printers and $M=19$ rounds for Case 1.}\label{FIG:res_1}
\end{figure*}

\paragraph{\textbf{Case 1: Full Fleet Participation}} The results for Case 1 where each of the $K=10$ printers is allowed to perform an experiment in each of the $M$ rounds are shown in Table~\ref{tab:res_2} and Fig. \ref{FIG:res_1}.

\begin{table}[ht]
\centering  
\caption{Average number of trials needed to find the optimal process parameters for Case 1: Full fleet participation.}
\label{tab:res_2}
\begin{tabular}{c c c}  
\toprule
\textbf{Printer No.} & \textbf{Collaborative} & \textbf{Non-collaborative} \\ 
\midrule
1  & 2  & 10\\
2  & 2  & 1 \\
3  & 2  & 8  \\
4  & 5  & 3  \\
5  & 16 & 18 \\
6  & 1 & 3 \\
7  & 2 & 7 \\
8  & 14 & 13 \\
9  & 11  & 18 \\
10 & 2 & 17 \\
\hline
Average & 5.7 & 9.8 \\
\bottomrule
\end{tabular}
\end{table}

Based on the results, several important insights can be drawn. First, as shown in Table~\ref{tab:res_2}, our approach achieves a 41.8\% reduction in the average number of trials required to find the optimal settings (from 9.8 to 5.7). This highlights the ability of printers to borrow strength from each other, thereby accelerating the discovery of optimal process parameters. Second, we observe that across nearly all printers, except for Printer 8, where results are comparable, collaboration consistently helps identify the optimal process parameters more quickly. This sheds light on the ability of collaboration to benefit all participants. 

Third, our approach achieves lower cumulative regret across most printers. This can be deduced from Fig.~\ref{FIG:res_1}, which shows that the terminal cumulative regret of the collaborative case is often lower than the non-collaborative case for most printers (namely, printers 3, 5, 6, 7, 8 and 9). This indicates that not only do we reach the optimal settings faster, but also that the decisions made along the way tend to be closer to optimal, compared to the non-collaborative baseline. Returning to the earlier example of defective parts, this implies that under a collaborative strategy, significantly fewer parts would need to be discarded—since each trial yields higher utility. Finally, the collaborative approach exhibits fewer and less erratic jumps in regret compared to the non-collaborative method, resulting in a noticeably smoother learning curve. While occasional jumps are expected, often reflecting strategic exploration of potentially better parameter settings, the reduced volatility highlights the stability and efficiency of our method. This smoother convergence suggests that our approach is better at leveraging shared structure across printers, enabling it to explore more intelligently and make consistently better decisions over time.

\begin{figure*}
	\centering
	\includegraphics[width=\textwidth]{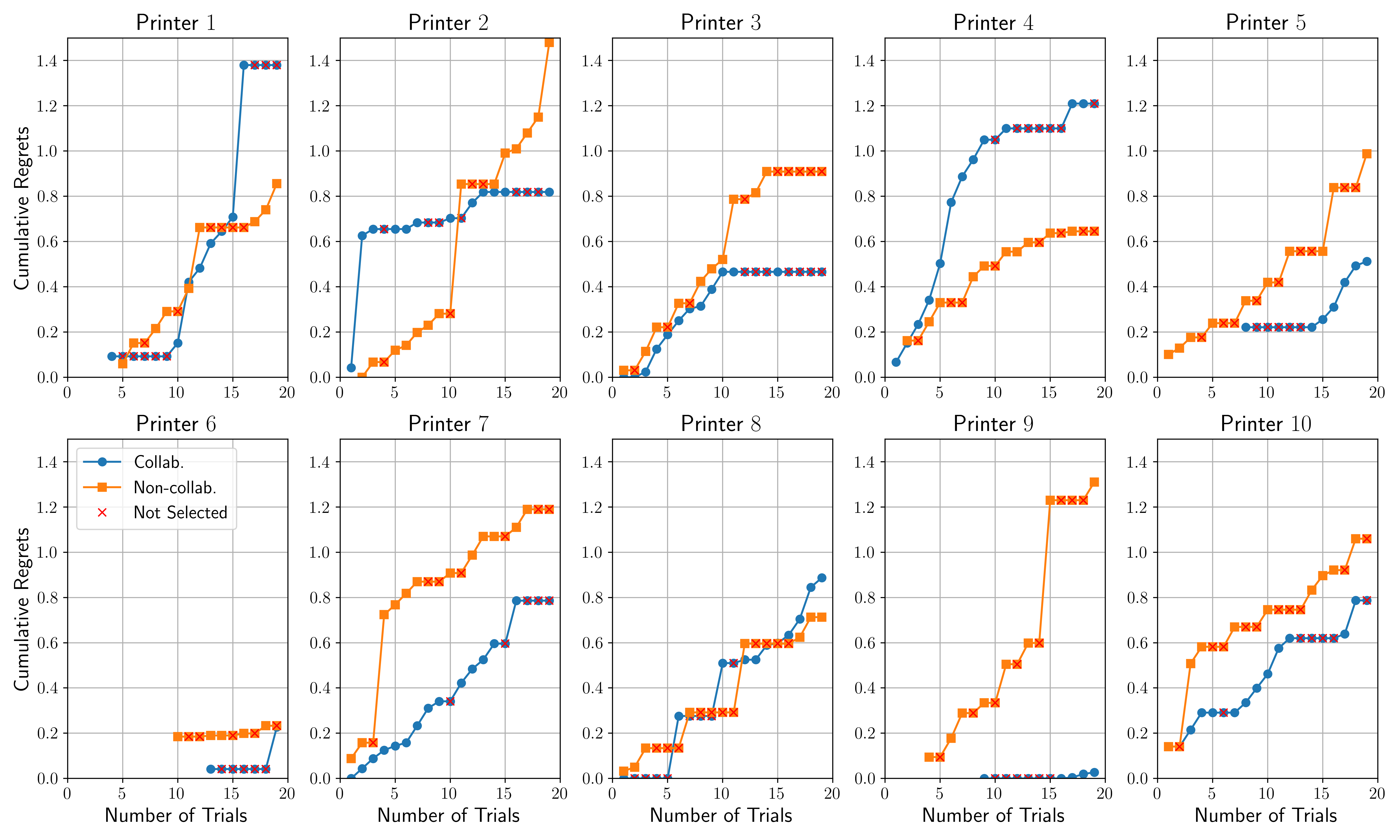}
	\caption{Cumulative regret of our proposed collaborative MC benchmarked against non-collaborative MC over $K=10$ printers and $M=19$ rounds for Case 2 ($c=5$). Each red cross denote an iteration $t$ where a printer is not selected for experimentation.
    }
	\label{FIG:sub_res_1}
\end{figure*}

\paragraph{\textbf{Case 2: Partial Fleet Participation}} The results for Case 2 where only $c=5$ printers are selected per round for experimentation are shown in Table~\ref{tab:sub_res_2} and Fig. \ref{FIG:sub_res_1}. Note that in Fig. \ref{FIG:sub_res_1}, many printers have missing regret values at some iterations as they are not selected in a specific round $t$. In addition, some printers failed to reach the true optimal process parameter setting within the total experimental budget in this case. Therefore, for both the baseline and our proposed method, we 
estimate the probability of failure on finding the true optimal setting at trial $t$ as $\widehat{P}(t)$ and the average number of trials as $\hat{\mu}$ using the Kaplan-Meier estimator \cite{Kaplan1958}, 
\begin{equation*}
    \widehat{P}(t)=\prod_{i:\:t_{i}\leq t}\left(1-\frac{g_{i}}{h_{i}}\right), \quad \hat{\mu}=\int_{0}^{M}\widehat{P}(t)dt,
\end{equation*}
with $t_{i}$ representing a specific round with at least one printer being able to reach the optimal setting, $g_{i}$ representing the number of printers reaching the optimal setting in round $t_{i}$, and $h_{i}$ representing the number of printers that failed to reach the optimal setting up to round $t_{i}$.

\begin{table}[ht]
\centering 
\caption{Average number of trial needed to find the optimal process parameters in Case 2, where only subset of $c=5$ printers are selected experimentation each round.}
\label{tab:sub_res_2}
\begin{tabular}{c c c}  
\toprule
\textbf{Printer No.} & \textbf{Collaborative} & \textbf{Non-collaborative} \\ 
\midrule
1  & \textgreater19  & \textgreater19  \\
2  & 5  & 2  \\
3  & 1  & \textgreater19  \\
4  & 18  & 12  \\
5  & 14 & 15 \\
6  & \textgreater19  & 14  \\
7  & 1  & 14  \\
8  & 1 & 19 \\
9  & 9 & \textgreater19 \\
10 & 5  & \textgreater19 \\
\hline
Average & 9.2 & 15.2\\
\bottomrule
\end{tabular}
\vspace{1ex} 
\end{table}

Once again, we observe from Table~\ref{tab:sub_res_2} that even with only half of the fleet participating in each round, the collaborative method yields a substantial average reduction of 39.5\% in the number of trials required to identify optimal parameters. This highlights the robustness of the collaborative approach under realistic operational constraints such as limited staffing, material availability, or restricted testing capacity. That said, the increased experimentation burden under partial participation is evident: the average number of iterations needed to reach the optimal configuration increased from 5.7 in Case 1 (full participation) to 9.2 in Case 2. This reflects the natural trade-off between resource availability and convergence speed. Besides that, the regret plots in Fig.~\ref{FIG:sub_res_1} demonstrate that collaborative learning continues to consistently achieve lower cumulative regret across most printers, even under constrained settings. This reaffirms the method's ability to effectively prioritize which machines to test and which parameters to explore, enabling more informed and efficient decisions throughout the experimentation process.

\section{Conclusions and Future Work} \label{Sec:conclusion}
This work introduces a novel machine learning-based collaborative recommender system for optimizing process parameters across fleets of networked manufacturing machines. By modeling the task as a sequential matrix completion problem, the proposed approach enables machines to collaboratively learn from shared data while tailoring recommendations to their individual characteristics. 

Using a 3D printing farm as a representative example, we demonstrate the system's effectiveness in a practical setting. The method achieves a 41.8\% reduction in the number of trials needed to find optimal process parameters under full fleet participation and a 39.5\% reduction under partial participation, compared to non-collaborative baselines. It also consistently delivers lower cumulative regret, reflecting more efficient decision-making and better interim performance throughout the optimization process.

These quantitative improvements have meaningful implications in the real world. In manufacturing environments with large machine fleets, the reduced trial count leads to lower operational overhead, faster deployment of optimal settings, and minimized material waste. The framework also demonstrates resilience under real-world constraints such as limited testing capacity. While demonstrated here with 3D printers, the methodology is broadly applicable to other manufacturing domains where process parameter tuning is critical. Future extensions will incorporate federated learning to support privacy-preserving optimization across geographically distributed systems.

\section*{Declaration of Generative AI and AI-assisted technologies in the writing process}

During the preparation of this work the authors used ChatGPT 4.0 in order to improve the grammar and sentence structures of the manuscript. After using this tool/service, the authors reviewed and edited the content as needed and take full responsibility for the content of the publication.

\section*{Acknowledgements}

This work was supported by a research grant from Cisco Systems, Inc.

\bibliographystyle{model1-num-names}

\bibliography{cas-refs}
\end{document}